\documentclass{article}
\usepackage{spconf,amsmath,epsfig}
\usepackage{url}
\usepackage{balance}
\usepackage{amsmath}
\usepackage{graphicx}
\graphicspath{{Figures/}}
\usepackage{pdfpages}
\usepackage{cite}
\usepackage{lipsum}
\usepackage{subfig} 
\usepackage[T1]{fontenc} 

\usepackage{array}
\usepackage{url}
\usepackage{float}

\usepackage{color}
\usepackage{wrapfig}
\usepackage{lipsum}
\usepackage[hidelinks]{hyperref}

\usepackage{multirow}

\usepackage{mathtools}

\usepackage{gensymb} 

\usepackage[font=small,labelfont=bf,tableposition=top]{caption}
\usepackage{amsmath}
\usepackage{amssymb}  
\usepackage{amsfonts}

\title{UCloudNet: A Residual U-Net with Deep Supervision for Cloud Image Segmentation}
\name{Yijie~Li$^{1}$, Hewei~Wang$^{1}$, Shaofan~Wang$^{2}$, Yee~Hui~Lee$^{4}$, Muhammad~Salman~Pathan$^{5}$, Soumyabrata~Dev$^{1,3}$\thanks{The ADAPT Centre for Digital Content Technology is funded under the SFI Research Centres Programme (Grant 13/RC/2106\_P2) and is co-funded under the European Regional Development Fund.}\thanks{Send correspondence to \mbox{S Dev, E-mail: soumyabrata.dev@ucd.ie.}}}
\address{
        $^{1}$School of Computer Science, University College Dublin, Dublin D04V1W8, Ireland \\
	$^{2}$Faculty of Information Technology, Beijing University of Technology, Beijing 100124, China \\ 
	$^{3}$The ADAPT SFI Research Centre, Dublin D04V1W8, Ireland \\
        $^{4}$School of Electrical and Electronic Engineering, Nanyang Technological University, Singapore \\
	$^{5}$Innovation Value Institute, Maynooth University, Ireland \\
}

\begin{document}
%
\maketitle
\begin{abstract}
Recent advancements in meteorology involve the use of ground-based sky cameras for cloud observation. Analyzing images from these cameras helps in calculating cloud coverage and understanding atmospheric phenomena. Traditionally, cloud image segmentation relied on conventional computer vision techniques. However, with the advent of deep learning, convolutional neural networks (CNNs) are increasingly applied for this purpose. Despite their effectiveness, CNNs often require many epochs to converge, posing challenges for real-time processing in sky camera systems. In this paper, we introduce a residual U-Net with deep supervision for cloud segmentation which provides better accuracy than previous approaches, and with less training consumption. By utilizing residual connection in encoders of UCloudNet, the feature extraction ability is further improved. In the spirit of reproducible research, the model code, dataset, and results of the experiments in this paper are available at: \url{https://github.com/Att100/UCloudNet}.
\end{abstract}

\section{Introduction}
Cloud information analysis plays a crucial role in the field of meteorological research, offering valuable insights into weather patterns and facilitating enhanced forecasting methods. As computer vision and machine learning advance, they have expanded into various interdisciplinary fields such as meteorology estimation \cite{dey2023nesnet,dev2019estimating, akrami2022graph} and weather variable prediction \cite{manandhar2019data, wang2021day, jain2024holistic}. While traditionally cloud imagery has been captured via meteorological satellites in near-earth orbit, the recent utilization of ground-based sky cameras ~\cite{jain2021extremely,dev2015design} has gained prominence due to their enhanced temporal and spatial resolution. With the advent of these cameras, several datasets of optical RGB images, such as  SWIMSEG \cite{dev2016color}, SWINSEG \cite{dev2017nighttime}, and SWINySEG \cite{dev2019cloudsegnet}. To better extract the cloud information from those images, the deep learning method with a fully convolution network (FCN) is widely used for cloud segmentation which consists of a series of encoders and decoders. However, this early design without an extra modified structure makes it hard to aggregate features from the first few layers. Additionally, training deep convolutional neural networks remains challenging without a shortcut skip connection.

In this paper, we propose the residual U-Net model with deep supervision for the cloud segmentation task, called UCloudNet. UCloudNet consists of a series of convolution blocks with residual connections which enhance the feature aggregation ability of the original U-Net by including more feature map fusion operations. The experiments prove that our proposed UCloudNet can achieve better performance with less training time and iterations than previous approaches. 

The main contributions of our work are threefold:
\begin{itemize}
	\item We propose UCloudNet, a novel U-Net based model enhanced with residual connections in encoder, significantly improving capability for feature extraction.
	\item We adopt deep supervision in our UCloudNet which substantially reduces the training time consumption.
	\item We conduct extensive experiments on three different benchmark datasets and validate the effectiveness of our proposed modules and training strategy. 
\end{itemize}

\begin{figure*}[htbp]
	\centering
	\includegraphics[width=7in]{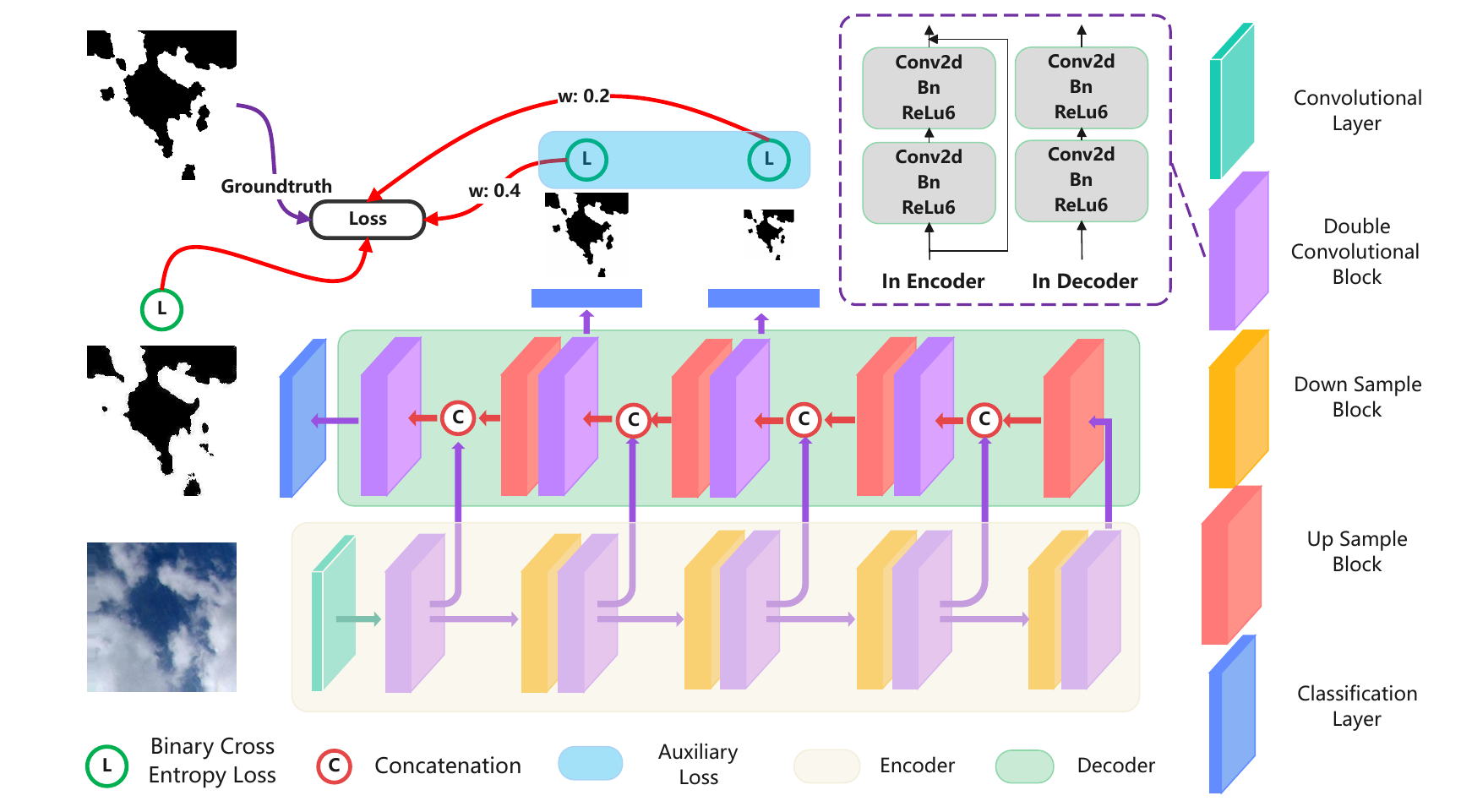}
	\caption{The architecture of the UCloudNet model. The procedure between the output of model and the segmentation mask has been omitted in this figure.}
	\vspace{-0.3cm}
	\label{fig:model-structure}
\end{figure*}

\begin{figure}[htbp]
	\centering
		\includegraphics[width=0.8in]{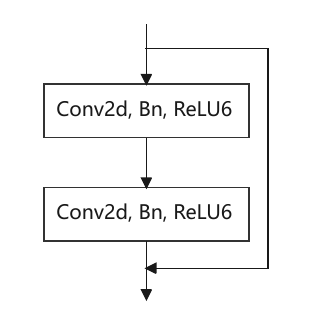}
		\includegraphics[width=0.8in]{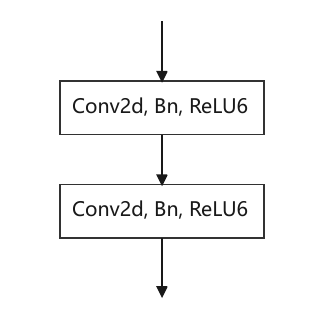}
		\includegraphics[width=0.8in]{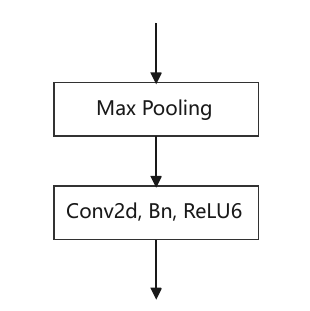}
		\includegraphics[width=0.8in]{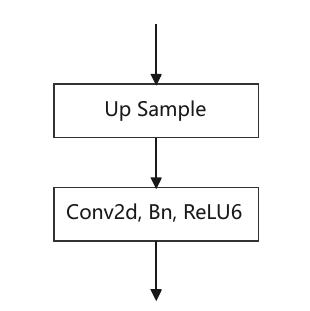}
	\caption{The structure of ‘Double Convolution Block’ in encoder, decoder, ‘Down Sample Block’, and ‘Up Sample Block’ (from left to right).}\label{fig:modules structure}
\end{figure}

\section{Related Works}
In addressing the challenge of segmenting sky/cloud images, a variety of methods have been developed. These methods are generally categorized into traditional computer vision techniques, as outlined in visual model related studies like~\cite{long2006retrieving, dev2014systematic, yang2010automatic}, and more recent deep learning approaches, as demonstrated in~\cite{dev2019cloudsegnet,dev2019multi} multi-label image segmentation. Traditional techniques often rely on color feature analysis, static convolution filters, and pixel gradients. For instance, Dev \textit{et al}. \cite{dev2014systematic} use principal component analysis (PCA) and fuzzy cluster to evaluate the color model aims to capture the greatest color variance between cloud and sky in 2014. While these methods are adept at capturing the general sky distribution, they tend to lose finer details, resulting in suboptimal segmentation accuracy and imprecise binary masks that don’t align well with the image boundaries. In contrast, the introduction of deep learning techniques has notably enhanced the quality of binary cloud mask generation. As for some learning-based computer vision tasks, CNN is a type of deep neural network widely utilized for analyzing 2D images~\cite{WANG2022102243, BATRA2022200039, WANG2021, WANGCAR2022} and attention mechanisms are also frequently employed to enhance visual feature representation~\cite{HuoAtten, Lin2024JointPT, Huo2023HumanorientedRL, xu2024mentor}. The encoder-decoder-based network is the most popular approach for segmentation and object detection-related tasks \cite{zhenqi23}, and Li \textit{et al}. \cite{li2023daanet} adopt a network with dual attention to perform salient object detection.  Dev \textit{et al}. \cite{dev2019cloudsegnet} CloudSegNet introduced an innovative approach in 2019, utilizing a standard Fully Convolutional Network (FCN) framework. This method compresses information into high-dimensional feature maps through downsampling, followed by upsampling procedures to refine segmentation results. This significantly improves boundary detail and overall segmentation accuracy. Additionally, Dev et al. \cite{dev2019multi} presented a novel multi-label sky/cloud segmentation strategy in 2019, utilizing a multi-class U-Net model to classify images into three categories, enabling more precise analysis via three-class segmentation.

\begin{figure*}[htbp]
	\centering
	\includegraphics[height=2.11in]{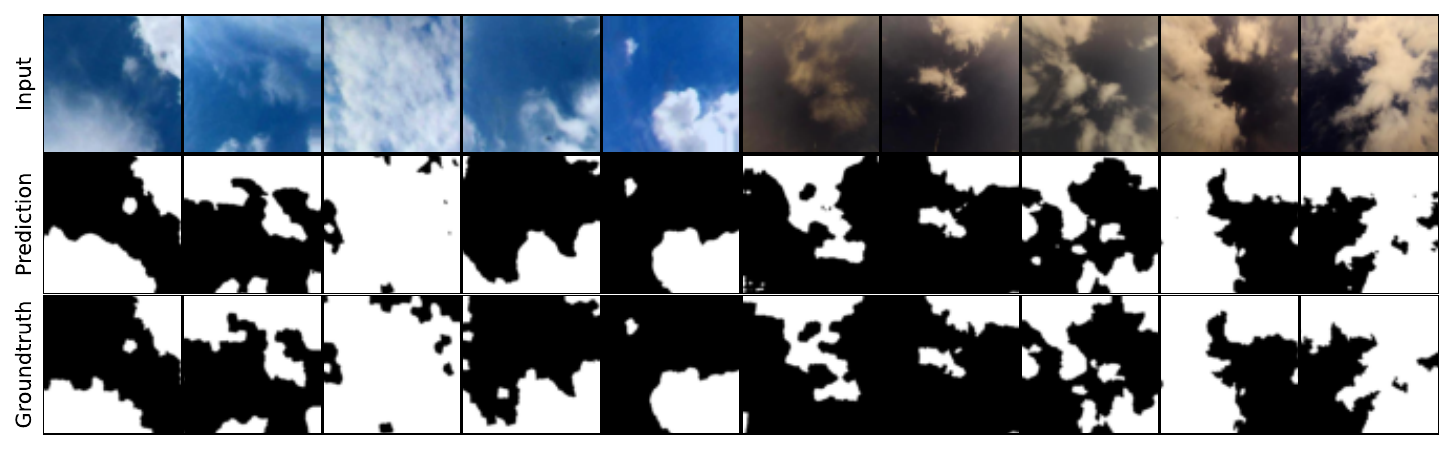}
	\caption{Results of cloud segmentation for day-time (1-6 columns) and night-time (7-12 columns).}\label{fig:Pred}
\end{figure*}

\section{Architecture}
\label{sec:format}

Our UCloudNet is based on the U-Net \cite{ronneberger2015u} structure which contains a series of decoders and encoders with channels concatenation in each stage. To compare with the original U-Net structure, we use a hyper-parameter $k$ to control the parameters amount and inspired by ResNet, we add residual connection in each convolution block in encoder which is helpful for training the deeper layers. As for the training strategy, we use deep supervision to support the training process. The architecture of our proposed model is shown in Fig.~\ref{fig:model-structure}. Our model contains a series of ‘Double Convolution’ blocks, ‘Down Sample’ blocks, and ‘Up Sample’ blocks. We explain these blocks in the following sections. 

\subsection{Double Convolution Block (DCB)}
Fig.~\ref{fig:modules structure} describes the structure of the double convolution block. 
The ‘Double Convolution Block’ contains two groups of layers which include an original 3x3 convolution layer, a ReLU6 activation layer, and a batch-normalization layer. This group of layers is called ‘BasicConv2d’ in our implementation. The structure of the DCB module is different in encoder and decoder, we only apply residual connection in the DCB in encoder while the structure of DCB in decoder is just a simple stack of `Conv-Bn-ReLu6' groups. The different designs of DCB can result from the short-cut concatenation in the U-Net structure. The U-Net structure already has a short connection between encoder and decoder, and therefore there is no need to apply a short-cut in the decoder again.

\subsection{Down Sample Block and Up Sample Block}
The ‘Down Sample Block (DSB)’ includes a Max-pooling layer and a ‘BasicConv2d’ block. The ‘Up Sample Block (UPB)’ is a combination of an Up-sample layer and a ‘BasicConv2d’ block. The UPB module receives the output of a previous DCB and the output of this block will be concatenated with the output of another DCB in encoder with the same feature map size. 

\subsection{Deep supervision}
The deep supervision training strategy can effectively support the training of deep neural networks and improve the regularization ability. In our implementation, we use two additional auxiliary loss branches to enable deep supervision. These two loss branches are located at stages with $1/2$ and $1/4$ output size, as shown at the top left of Fig.~\ref{fig:model-structure}. In order to perform the binary cross-entropy loss, we interpolate the target map into $1/2$ and $1/4$ of the original size. Additionally, a group of weight values is considered in our total loss so that the significance of a high-resolution prediction map is larger than a low-resolution prediction map.

\subsection{Parameters control and loss function}
The number of parameters of our model can be controlled by a hyper-parameter $k$. The number of the filters of DCB in encoder can be specified by $k * 2^{s}$ while the configuration of DCB in decoder is $k * 2^{3-s}$. The filters of the convolution layer in the "Down Sample" block can be calculated by $k * 2^{s+1}$ and the filter number of the "Up Sample" block is $k * 2^{4-s}$, while $s=0,1,2,3$. 

We use binary cross entropy as the loss function and the total loss function can be represented as follows:

\vspace{-0.23in}
\begin{equation}
	L(p, y) = - \frac{1}{N} * \sum_{i=0}^{N} y_{i} * \log{p_{i}} + (1 - y_{i}) * \log{(1-p_{i})}
\end{equation} 
\vspace{-0.12in}
\begin{equation}
	L_{total}(p, y) = L(p, y) + 0.4*L(p_{2}, y_{2}) + 0.2*L(p_{4}, y_{4})
\end{equation}

\begin{table*}[htbp]
	\centering
	\caption{Comparison of UCloudNet with other cloud segmentation methods} \label{performances}
	\scalebox{1.1}{
 \small
	\begin{tabular}{c||r|cccc}
		\bf Dataset          &\bf Method          & \bf Precision & \bf Recall & \bf F-measure & \bf Error-rate \\
		\hline
		SWINySEG (day)        & 	Dev et al. 2014 \cite{dev2014systematic}& 0.89          & 0.92       & 0.89          & 0.09           \\
		(augmented SWIMSEG)   &     	Long et al. \cite{long2006retrieving}& 0.89          & 0.82       & 0.81          & 0.14           \\
							  &      	  Li et al. \cite{li2011hybrid}& 0.81          & \bf 0.97   & 0.86          & 0.12           \\
		                      &     	CloudSegNet \cite{dev2019cloudsegnet}& 0.92          & 0.88       & 0.89          & 0.07           \\
		                      &    	   Souza et al. \cite{souza2006simple}& \bf 0.99      & 0.53       & 0.63          & 0.18           \\ 
		                      &   UCloudNet(k=2)+aux+lr-decay& 0.90& 0.93       & 0.92          & 0.07           \\
		                      &   UCloudNet(k=4)+aux+lr-decay& 0.92& 0.94       & \bf 0.93      & \bf 0.06       \\ 
		\hline
		SWINySEG (night)      &     Dev et al. 2017 \cite{dev2017nighttime}&     0.94      & 0.74       &     0.82      &     0.13       \\
		(augmented SWINSEG)   &    Yang et al. 2009 \cite{yang2009image}& \bf 0.98      & 0.65       &     0.76      &     0.16       \\
		                      &    Yang et al. 2010 \cite{yang2010automatic}&     0.73      & 0.33       &     0.41      &     0.37       \\
		                      &        Gacal et al. \cite{gacal2016ground}&     0.48      & \bf 0.99   &     0.62      &     0.50       \\
		                      &         CloudSegNet \cite{dev2019cloudsegnet}&     0.88      & 0.91       &     0.89      &     0.08       \\
		                      &   UCloudNet(k=2)+aux+lr-decay& 0.92& 0.94       &     0.93      &     0.06       \\
		                      &   UCloudNet(k=4)+aux+lr-decay& 0.95& 0.95       & \bf 0.95      & \bf 0.04       \\
		\hline
		SWINySEG (day+night)  &         CloudSegNet \cite{dev2019cloudsegnet}& 0.92          & 0.87       & 0.89          & 0.08           \\
		                      &   UCloudNet(k=2)+aux+lr-decay& 0.90& 0.92       & 0.91          & 0.08           \\
		                      &      UCloudNet(k=4)& 0.92          & 0.90       & 0.91          & 0.08           \\
		                      &       UCloudNet(k=4)+lr-decay& 0.91& 0.94       & 0.92          & 0.07           \\
		                      &   UCloudNet(k=4)+aux+lr-decay& \bf 0.92& \bf 0.94       & \bf 0.93          & \bf 0.06           \\
	\end{tabular}
	}
	\vspace{-0.5cm}
\end{table*}

\section{Experiments \& Results}
We conduct experiments under different configurations, including the size of the model (controlled by $k$), usage of learning rate decay, and the deep supervision strategy.

\subsection{Dataset and Training Configurations}
The cloud segmentation dataset used in our study was sourced from the Singapore Whole Sky Nychthemeron Image SEGmentation Database (SWINySEG), including a total of $6078$ daytime and $690$ nighttime cloud images. Our proposed UCloudNet was developed using the PaddlePaddle framework and trained on a single NVIDIA Tesla V100-SXM2 16GB GPU. We partitioned the dataset into training and testing sets with an 8:2 split, and trained on three subsets: daytime, nighttime, and the complete SWINySEG dataset. The training was conducted with a batch size of 16 for a total of 100 epochs. We use Adam optimizer, employed with an initial learning rate of 0.001. We incorporated an exponential decay in the learning rate, with a gamma value of 0.95 applied after each epoch.

\subsection{Metrics}
We assessed the effectiveness of our model using four commonly adopted metrics: precision, recall, F-measure, and error rate. The F-measure is represented by  $\frac{2 \times Precision \times Recall}{Precision + Recall}$. Precision is defined as $\frac{TP}{TP+FP}$, while recall is $\frac{TP}{TP+FN}$, and the error rate is calculated using $\frac{FP+FN}{P+N}$. We employed the PR curve across 256 different thresholds.

\subsection{Quantitative Analysis}
Quantitative evaluation results of our methods are shown in Table 1, which shows the precision, recall, F-measure, and error-rate of our proposed UCloudNet with other methods on different data sets. On the full SWINySEG data set, UCloudNet (k=4) with deep supervision and learning rate decay has the best performance on all the four metrics while UCloudNet (k=4) with only learning rate decay has the second best performance which can prove that deep supervision with auxiliary loss can improve our model performance. On day-time images, the overall performance is better than others. As for the night-time images, our model has the lowest error rate. 

\begin{figure}[htbp]
	\centering
	\includegraphics[height=2.16in]{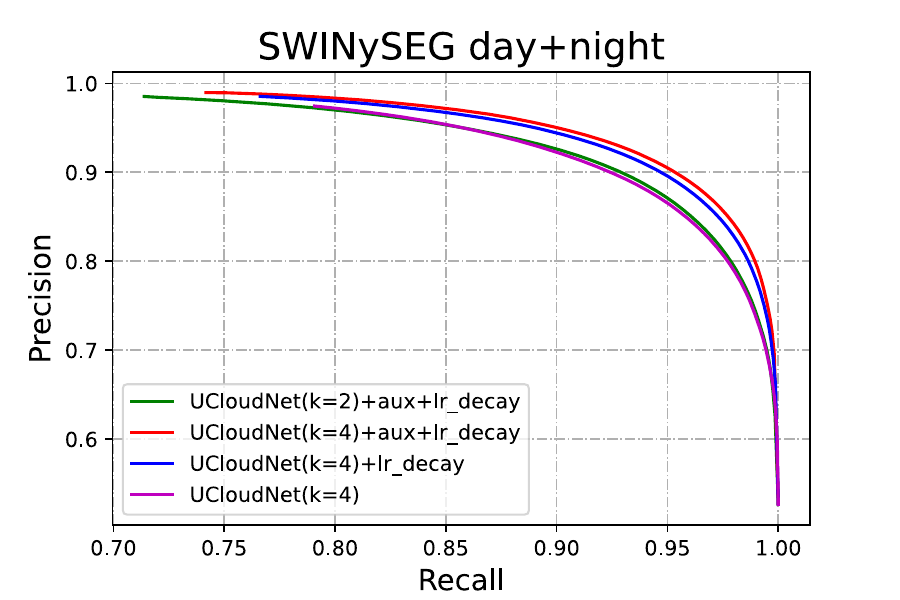}
    	\caption{PR curve of UCloudNet with different training configurations on full SWINySEG ground-based cloud segmentation dataset.}
	\label{fig:pr_curve}
	\vspace{-0.3cm}
\end{figure}

\subsection{Qualitative Analysis}
Qualitatively, we use several day-time and night-time images as our evaluation samples. We use these images as input and perform a threshold with p=0.5 on the sigmoid output, shown in Fig.~\ref{fig:Pred}. We observed our binary prediction maps and compared them with input images and ground truths, it is proved that our proposed model can accurately extract the overall feature and generate binary prediction correctly. However, there still exists some negative phenomenons, for example, the lack of accuracy on segmentation on edges. Besides, we evaluate our proposed model with the PR curve, shown in Fig.~\ref{fig:pr_curve}. In order to receive clearer and more precise results, we set 256 thresholds to generate the curve, the area under the curve is bigger, and the overall performance of the model is better. 

\begin{figure}[H]
	\centering
	\includegraphics[height=2.4in]{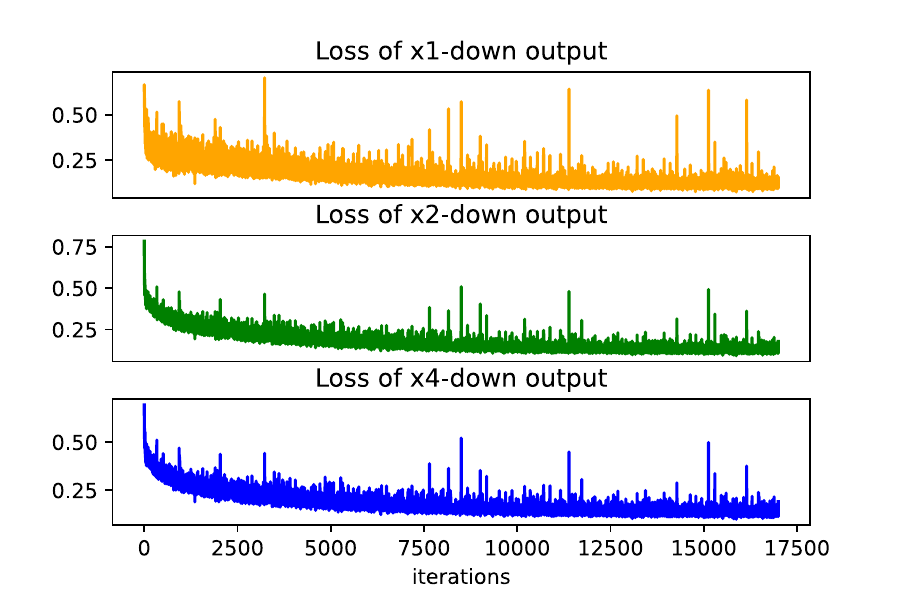}
	\caption{Loss curve of the final output and auxiliary outputs.}
	\label{fig:Loss-curve}
\end{figure}

Additionally, we evaluate the training status of our proposed model qualitatively by observing curves of final loss together with auxiliary loss branches, shown in Fig.~\ref{fig:Loss-curve}. The loss curves show that the loss of the final output converges much faster than the loss of x2-down-sample loss and x4-down-sample loss in the early 2500-iterations, but the tendency tends to be the same after it, which can prove that the auxiliary loss is helpful to the training of deep convolution neural network at the beginning. From another point of view, the loss can achieve a low and stable level in less than 10,000 iterations, and then keep dropping slowly until it finally converges which proves that the speed and ability of fitting of our proposed model is considerable. 

\section{Conclusion}
In this paper, we introduce a residual U-Net with deep supervision for cloud-sky segmentation. We train our model with different configurations on various splits of SWINySEG dataset. Our proposed method achieves better performance as compared to the other methods and our experiments prove that deep supervision with auxiliary loss can gain better performance. Additionally, our model only needs less than $17500$ iterations ($100$ epochs with batch-size $16$) to converge which can significantly save the training consumption, as compared to other deep learning methods. For the future, we aim to develop more lightweight networks for tasks like multi-class pixel-wise classification and cloud depth estimation. building on our existing approach to enhance accuracy in various tasks while also increasing the speed of inference.


\bibliographystyle{IEEEtran.bst}
\bibliography{strings,refs}

\end{document}